\documentclass[lettersize,journal]{IEEEtran}
\usepackage{amsmath,amsfonts}
\usepackage{algorithmic}
\usepackage{algorithm}
\usepackage{array}
\usepackage[table]{xcolor}
\usepackage{textcomp}
\usepackage{stfloats}
\usepackage{url}
\usepackage{bm}
\usepackage{verbatim}
\usepackage{graphicx}
\usepackage{cite}
\makeatletter
\let\c@lofdepth\relax
\let\c@lotdepth\relax
\makeatother
\usepackage{subfigure}
\usepackage{amssymb}
\usepackage{multirow}
\usepackage{color}
\usepackage{booktabs} 
\usepackage{makecell}
\usepackage{float}
\usepackage{hyperref}

\hypersetup{hidelinks,
	colorlinks=true,
	allcolors=blue,
	pdfstartview=Fit,
	breaklinks=true}
\begin{document}

\title{Answering Diverse Questions via Text Attached with Key Audio-Visual Clues}

\author{Qilang Ye, Zitong Yu, and Xin Liu
\thanks{This work was supported in part by the National Natural Science Foundation of China under Grant 62306061. (Corresponding authors: Zitong Yu)}
\thanks{Q. Ye is with the School of Computing and Information Technology, Great Bay University, Dongguan 523000, China, and also with the School of Computer Science and Engineering, Chongqing University of Technology, Chongqing 401300, China (e-mail: rikeilong@stu.cqut.edu.cn).}
\thanks{Z. Yu is with the School of Computing and Information Technology, Great Bay University, Dongguan 523000, China. (email: zitong.yu@ieee.org)}
\thanks{X. Liu is with the Computer Vision and Pattern Recognition Laboratory, Lappeenranta-Lahti University of Technology LUT, 53850 Lappeenranta, Finland (e-mail: linuxsino@gmail.com)}
}


\maketitle

\begin{abstract}
Audio-visual question answering (AVQA) requires reference to video content and auditory information, followed by correlating the question to predict the most precise answer.  Although mining deeper layers of audio-visual information to interact with questions facilitates the multimodal fusion process, the redundancy of audio-visual parameters tends to reduce the generalization of the inference engine to multiple question-answer pairs in a single video. Indeed, the natural heterogeneous relationship between audiovisuals and text makes the perfect fusion challenging, to prevent high-level audio-visual semantics from weakening the network's adaptability to diverse question types, we propose a framework for performing mutual correlation distillation (MCD) to aid question inference. MCD is divided into three main steps: 1) firstly, the residual structure is utilized to enhance the audio-visual soft associations based on self-attention, then key local audio-visual features relevant to the question context are captured hierarchically by shared aggregators and coupled in the form of clues with specific question vectors. 2) Secondly, knowledge distillation is enforced to align audio-visual-text pairs in a shared latent space to narrow the cross-modal semantic gap. 3) And finally, the audio-visual dependencies are decoupled by discarding the decision-level integrations. We evaluate the proposed method on two publicly available datasets containing multiple question-and-answer pairs, i.e., Music-AVQA and AVQA. Experiments show that our method outperforms other state-of-the-art methods, and one interesting finding behind is that removing deep audio-visual features during inference can effectively mitigate overfitting. The source code is released at \href{github}{\textcolor{black}{http://github.com/rikeilong/MCD-forAVQA}}.
\end{abstract}


\section{Introduction}
\IEEEPARstart{T}HE Audio-visual Question Answering (AVQA) task \cite{musicavqa}, \cite{avqa}, \cite{pianoavqa} has driven significant advances in the field of multimodal learning and has become one of the indispensable components of human-computer interaction. Typical AVQA tasks include audio-visual representation learning as well as derivation based on multiple questions contained in a long video in order to obtain the most appropriate one of the candidate's answers. Vision and auditory senses are the main mediums for disseminating information in reality and are crucial for guiding questions to find answers from videos, however, when converted into machine-understandable semantics pose many challenges to the multimodal reasoning process.

Many works \cite{r4}, \cite{r6}, \cite{r7}, \cite{r8}, \cite{r9}, \cite{r10}, \cite{r11} focus on the audio-visual association, including audio-visual cross-attention, audio-visual contrastive learning \cite{avcl}, \cite{audioclip}, and masked modeling of audio-visual \cite{avmae}, \cite{clmae}. Although comprehensive scenario information facilitates reasoning, approaches focusing on exploring global audio-visual joint perception ignore the severe information redundancy generated by complex multimodal interactions in long videos, which does not adapt to diverse question types. A regular process of answering questions should be human-like, when we receive the question ``\textit{What is the first instrument that comes in}", we expect to locate important periods in the audio through the question keyword ``\textit{first}", identify corresponding spatial locations on the video through the keyword ``\textit{instrument}", followed by guiding the question to find out the most correct answer via these ``clues". Indeed, we do not expect visual and audio to join inference and argue that audio-visual-text late fusion may contribute to single-sample recognition but limits the generalization of its network. The reasons are as follows: a single video can generate multiple question-answer pairs, and the same question content can exist in multiple videos, so an individual audio-visual scenario has limitations in deriving answers corresponding to multiple questions. Second, the high-level semantics of audiovisuals are ambiguous with natural language, and their inconsistency leads to a susceptibility to generate noise when performing mutual concatenate and element-wise add.

\begin{figure} \centering    
\subfigure[Mutual correlation process] {
\includegraphics[scale=0.32]{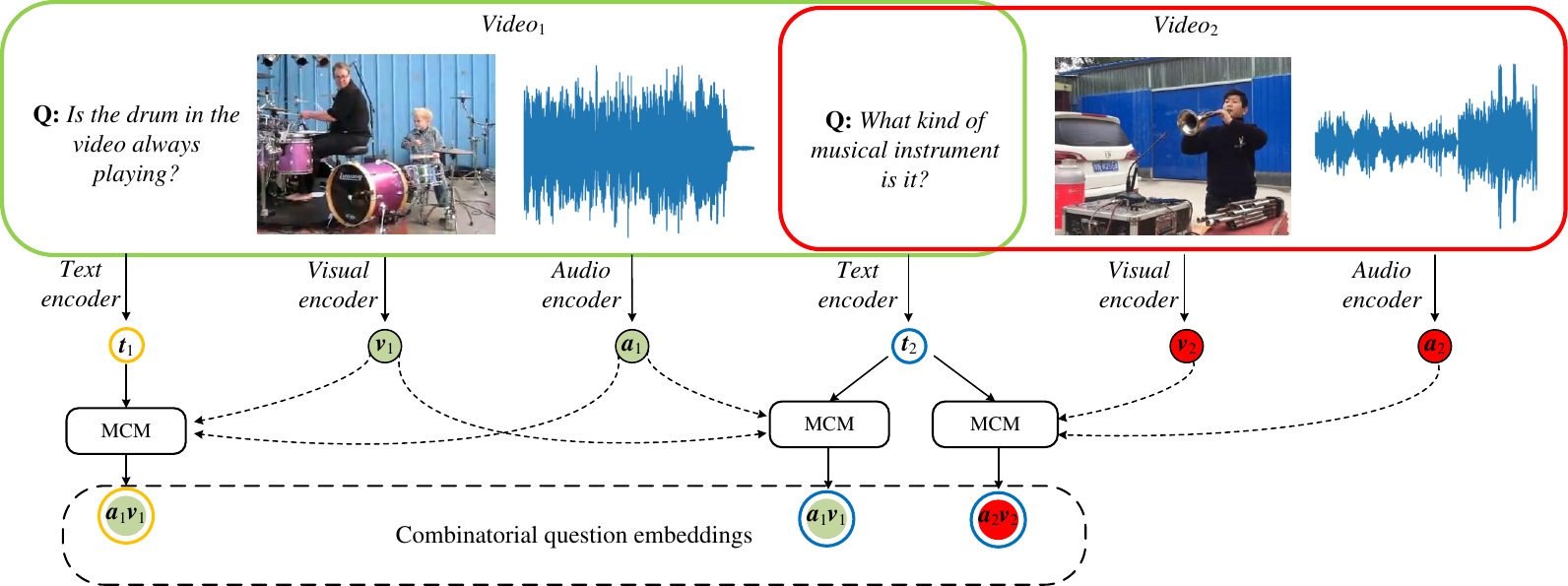}
}

\subfigure[Knowledge Distillation via Contrastive Learning] { 
\includegraphics[scale=0.45]{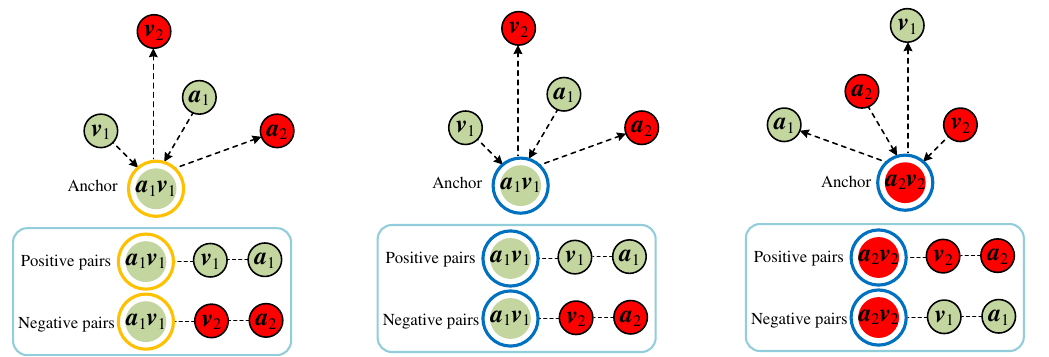}
}    

\caption{An illustration of our proposed mutual correlation distillation guidance. MCM denotes the mutual correlation module. The question content generated in multiple videos may be identical, thus in proposing the concept of knowledge distillation via contrastive learning, we first guide the question through the audio-visual to distinguish differences in the same question across samples. (a) is a simplified generation process of combinatorial question embedding. (b) is our proposal to alleviate semantic ambiguity between cross-modalities by approximating positive pairs and separating negative pairs in a contrastive way.}     
\label{fig:fig1} 
\end{figure}

To accommodate multiple types of question reasoning, we designed a multimodal interconnection framework called the Mutual Correlation Distillation (MCD). As shown in Fig. \ref{fig:fig1}, the network mainly proposes two steps to equip the question with the ability to perceive the answer: first, the visual, auditory, and textual information is transformed into machine-understandable representations via basic encoders, and then uniformly fed into the Mutual Correlation Module (MCM) proposed in this paper, which adaptively generates unique clues describing the scene using the captured coordinated audio-visual information and intersects them with the original question features to form a specific problem representation named combinatorial question embeddings. Secondly, we propose a Semantic Approximation (SA) method, mainly used to narrow the semantic gap between the combinatorial question embeddings and the original audio-visual in a contrastive learning way. Specifically, combinatorial question embeddings will serve as anchors to draw closer to audio-visual features within the sample and pull away from audio-visual features in other samples to enable cross-modal knowledge transfer. Our idea is simple, aiming to use only combinatorial question embeddings to predict the goal in order to minimize the presence of redundant and bias-prone parameters in the inference phase. In summary, the main contributions of this work are summarized as follows:

\begin{enumerate}
\item[1)]We propose a way of answering questions from a human perspective that no need for global audio-visual features to be incorporated into decision-level fusion, reducing the risk of overfitting due to redundant audio-visual parameters. Compared to previous work, this novel approach is more applicable to multiple question-answer pairs tasks.
\item[2)]We design a mutual correlation module that excels at perceiving audio-visual regions of interest and progressively guides the question features to comprehend the video content. It needs no exhaustive semantics of raw details and prevents the overloading of parameters that can damage the network generalization performance.
\item[3)]Since typical hard attention tends to allow partial modal information missing, we propose a soft association mechanism that skillfully applies residual structure to audio-visual self-attention to enhance cross-modal transmission while consolidating its information.
\item[4)]We outperform other state-of-the-art methods on the Music-AVQA and AVQA datasets, and conduct extensive experiments to demonstrate the strong generalization of MCD to common backbone networks.
\end{enumerate}

\section{Related Works} 
\subsection{Audio-Visual Question Answering}
\subsubsection{Video Question Answering Datasets}
Starting from visual questions answered on images \cite{r18}, several research tasks \cite{r19}, \cite{r20}, \cite{r21}, \cite{r22}, \cite{r23}, \cite{r24}, \cite{r25}, \cite{sutdtraffic} on video questions answered in video scenarios have been derived, which involve complex dynamic scenarios such as movies \cite{r19}, TV programs \cite{r23}, \cite{r25}, traffic \cite{sutdtraffic}, web GIFs \cite{r20}, animations \cite{r22}, and music. For video datasets, in addition to the image information of each frame in the video, audio features or subtitles text becomes another important medium for content dissemination. Among them, the \cite{r23} and \cite{r24} datasets collect video data in real environments that contain complex audio-visual pairs, which generate a large amount of redundant information prone to misclassification of the discriminative model. The \cite{musicavqa} and \cite{avqa} are cleaned data containing a large number of question-answer pairs and natural sounds, requiring the researcher to focus on the question content to perform a more complex inference process, and our work is based on the above two datasets as well.

\subsubsection{Video Question Answering Methods}
Many researchers have emphasized the association between visual and audio modalities to achieve effective audio-visual scene understanding. Early approaches focused on sound source localization \cite{r27}, \cite{r28}, \cite{r29}, \cite{r30}, \cite{r31}, \cite{r32}, or separate visuals \cite{depth-aware}, \cite{locans}, \cite{recover}, \cite{dual}, \cite{inter}, while these approaches can help in unimodal comprehension, they are so independent that they lose the ability to unite modalities. By integrating audiovisual information, existing work can improve performance by mining richer scene information in videos and extracting rich annotations in fine-grained audiovisuals. However, they focus on the perception of audiovisual objects in the video and lack attention to textual objects. Specifically, Yu \textit{et al.} \cite{long-term} argues that combining long-term audio-visual pairs produces redundant information that interferes with perceptual abilities, and therefore uses a Top-k-like idea of filtering segments that are irrelevant to the question, which, similar to \cite{pstp-net}, is effective but there is a high risk of losing useful information. Liu \textit{et al.} \cite{semantic} suggests understanding features at the fine-grained level of sentences, but for AVQA, the question itself contains no information worth exploring. Lin \textit{et al.} \cite{lavish} proposes an excellent dual-stream adaptive perception framework that dynamically learns interaction information from shallow audio-visual features, but general machines have difficulty with the huge amount of computational information that comes from the rich audio and images in the original video. Li \textit{et al.} \cite{musicavqa}, \cite{pstp-net} employ textual as a driver to establish spatial and temporal correlations between audiovisuals, as well as the combination of all modalities through concatenation. Yang \textit{et al.} \cite{avqa} proposes a hierarchical audio-visual fusing module, which performs through the fusion of features generated from audio, visual, and textual patterns at different stages. Although constructing perceptions between audiovisuals from multiple perspectives may provide an advantage in feature-level fusion, the natural heterogeneity across domains prevents decision-level fusion from being optimal. In terms of the AVQA task, utilizing potential audio-visual features to provide clues to predict the answer is more convincing than late fusion.

Moreover, recent approaches \cite{r37}, \cite{r38}, \cite{r39},\cite{r40}, \cite{r41}, \cite{r42}, \cite{r43} propose to utilize the expressive power of large-scale extra training data to solve AVQA tasks. However, training a large audiovisual-textual model from scratch takes a significant amount of time and the cost of collecting a video question answering dataset is prohibitive, therefore, we attempt to search for a way to still efficiently and accurately deduce the answer without the help of extra data, and MCD seems to be a promising solution, which focuses on understanding the question, with audiovisuals being only a tool to aid in comprehension. 

\begin{figure*}
	\centering
		\includegraphics[scale=0.6]{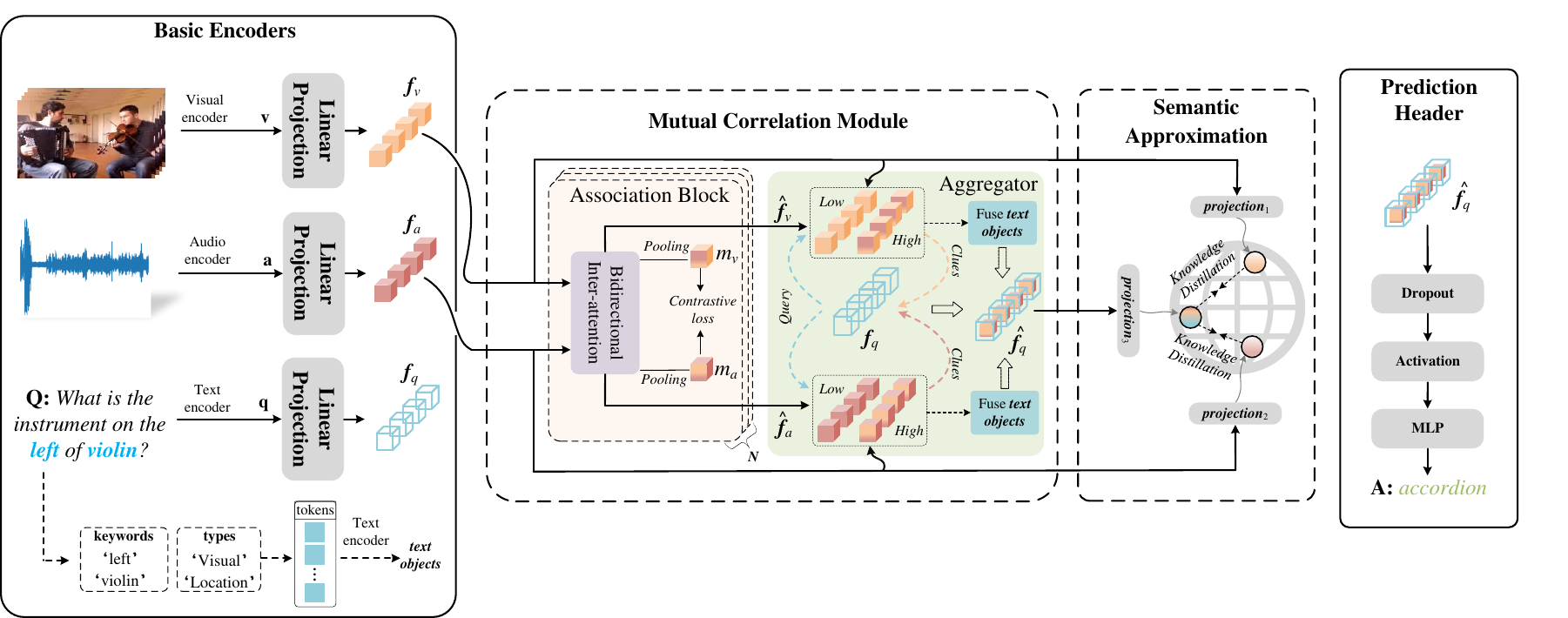}
	\caption{An overview of our MCD, where the dotted lines are the main contributions. The features $\bm{v}$, $\bm{a}$, and $\bm{q}$ are first obtained by the basic encoders, and to focus on the fine-grained semantic information in the sentences, we also combine the keywords with the question types to form text objects. Further, the audio-visual embeddings $f_{\bm{v}}$ and $f_{\bm{a}}$ will go through enhanced self-attention in $N$ association blocks to get the advanced audio-visual embeddings $\hat{f_{\bm{v}}}$ and $\hat{f_{\bm{a}}}$ (here we learn the audio-visual contrasts at the attention layer to emphasize coordination), which are then fed into the aggregator along with the question features $f_{\bm{q}}$ to generate the combinatorial question embeddings $\hat{f_{\bm{q}}}$, and it is worth noting that we add an additional optional branch to fuse the audio-visual embeddings with the text objects. In order for the combinatorial question embeddings to further learn the audio-visual knowledge, we propose to distill the knowledge in a shared latent space. Finally, the combinatorial question embeddings will be used to infer answers. }
	\label{fig:fig2}
\end{figure*}

\subsection{Transformer-based Cross-attention}
Since the great success of Transformer \cite{transformer} in natural language processing, a variety of Transformer-based visual tasks \cite{vit}, have also been rapidly developed. Benefiting from its excellent self-attention mechanism, associations between feature patches are computed by query to focus on local information. To extend attention across modalities, Chen \textit{et al.} \cite{cross-att} proposed a Transformer-based Cross-modal Attention mechanism, which flexibly combines multiple modalities in the form of queries to interlearn the association between modalities. Moreover, Yun \textit{et al.} \cite{pianoavqa} uses a Transformer-based cross-attention mechanism to cross-learn textual audio-visual joints, which its thoughts have informed our work. However, previous cross-modal attention methods are not effective for transferring information between heterogeneous data, the heterogeneous relationship between audiovisuals impairs its localization ability, and tokens as queries lose their information. Therefore, in this paper, we propose a novel soft cross-attention operation as a way to mitigate the information loss problem.


\section{Proposed Method}
In this section, we will introduce the implementation process of the mutual correlation distillation (MCD) framework. The overall architecture is shown in Fig. \ref{fig:fig2}, MCD consists of four panels: encoders for acquiring basic audio-visual and question features, MCM for adaptive generation of combinatorial question embeddings attached key audio-visual clues, SA for aligning combinatorial questions embeddings with original audio-visual, and a prediction header for inference the most appropriate answer.

\begin{figure*}
\centering    	
\subfigure[Details of our cross-attention.] {
\includegraphics[width=5.5cm]{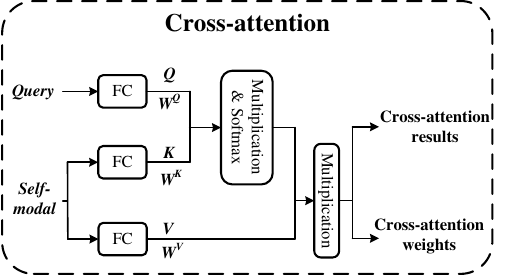}  		 
}    	 
\quad    	 
\hspace{2cm}\subfigure[Details of text obejects.] {
\includegraphics[width=5.5cm]{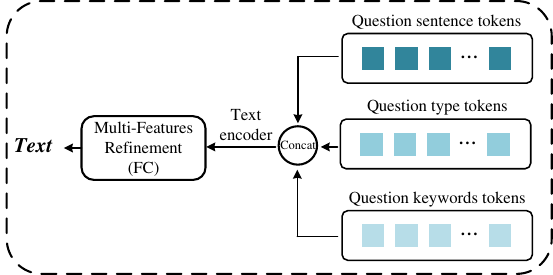}    		 
}    	  
\quad     	  
\subfigure[Details of association block and multimodal correlation process.] {
\includegraphics[scale=0.5]{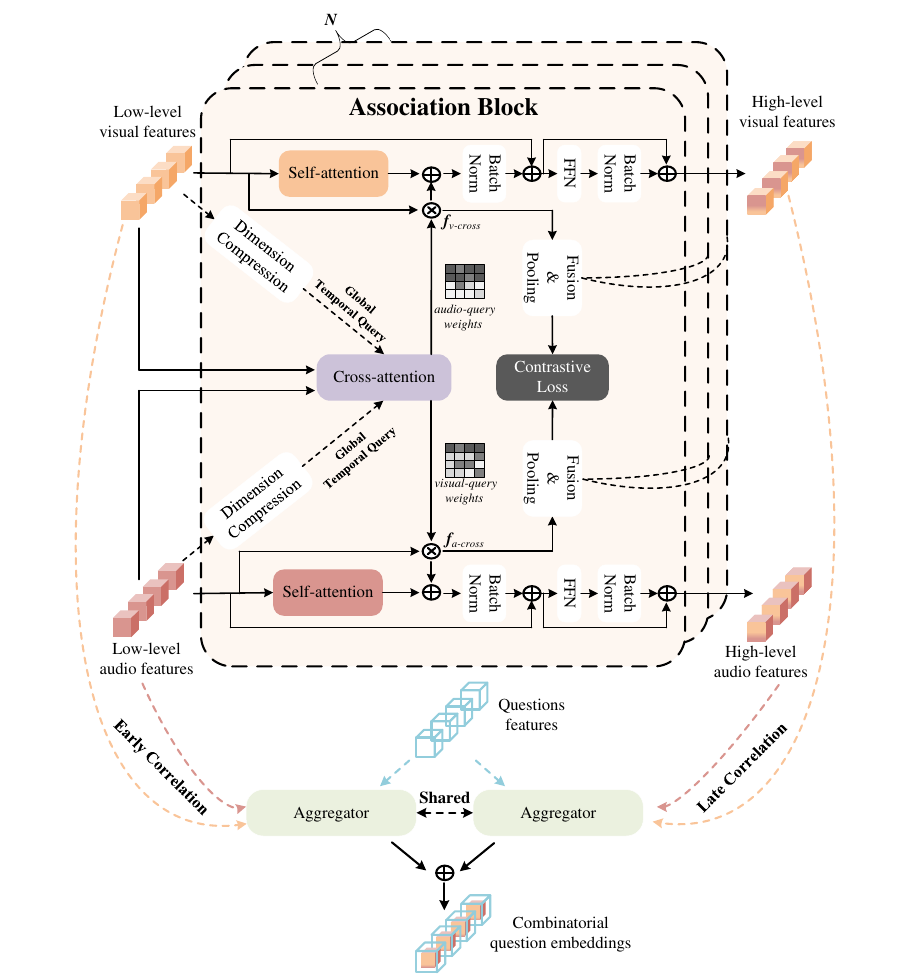}  	 	
}        	
\quad        	
\subfigure[Internal structure of the aggregator.] { 
\raisebox{0.1\height}{\includegraphics[scale=0.5]{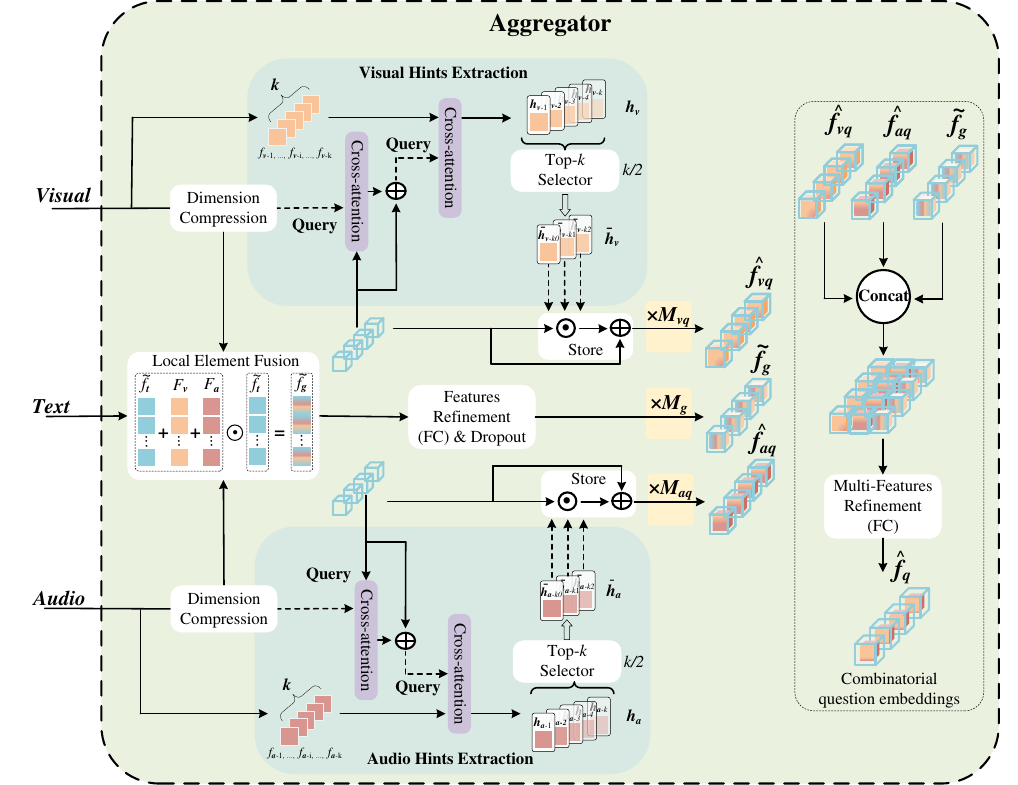}}   	
 }    
\caption{An overview of the Mutual Correlation Module (MCM). (a) proposes a Transformer-based cross-attention mechanism, (b) shows constituent elements of the text object, (c) demonstrates the multimodal correlation process, where the early correlation and late correlation are to ensure that both shallow independent audio-visual features and fine-grained audio-visual interaction features are learned in a balanced way, (d) demonstrates the internal structure of the aggregator. Specifically, the input audio-visual embeddings will first go through soft association to attach a small volume of information about each other to yield a high-level audio-visual embedding, followed by the low-high level audio-visual embeddings and the question features into the shared aggregator respectively, which additionally inputs text objects and finally outputs the combinatorial question embeddings.}      
\label{fig:fig3}	
\end{figure*} 

\subsection{Basic Encoders}
\cite{musicavqa}, as well as \cite{avqa} datasets, propose the use of pre-trained networks (e.g., ResNet \cite{resnet}, Vggish \cite{vggish}, CLIP \cite{r41}, etc.) to extract the intermediate layer audio-visual-text features. The basic encoders in this paper use similar pre-trained networks of current state-of-the-art methods for fair comparison, we also replace other backbone networks for generalization testing in the ablation experiments. 

\subsubsection{Audio-Visual-Text Inputs}
For \textbf{visualization}, the raw video is collected at the framerate of 60 fps by sampling 1 frame every second, after which it is passed through a visual encoder to obtain a visual intermediate representation ${\bm{v} \in {\mathbb{R}^{60 \times C_{\bm{v}} \times H \times W}}}$, where $C_{\bm{v}}$ denotes the number of image channels. For \textbf{auditory}, it can extract the sound from the raw video in MP3 or WAV format and convert it to Melspectrograms, which can be viewed as 2D images with width in frequency (128) and length (1024) in time. Equally, the spectrogram is extracted by the audio encoder to yield audio representation ${\bm{a} \in {\mathbb{R}^{60 \times C_{\bm{a}}}}}$, where $C_{\bm{a}}$ denotes the number of spectrogram channels. For \textbf{text}, we standardize the maximum length of each question and tokenization into word vectors, then encoding to $\bm{q} \in {\mathbb{R}^{L_q}}$, where $L_q$ denotes the max length of words.

\subsubsection{Audio-Visual-Text Embeddings}
We employ linear projection to flatten the frames, the spectrogram, and the word vector into unified feature embeddings. Specifically, the $L$ and $W$ of the visual image are performed with the $avgpool$ operation,  Subsequently, after the linear projection, the audio-visual-text is uniformly projected as visual embeddings ${f_{\bm{v}} \in {\mathbb{R}^{L_v \times D{\bm{v}}}}}$, audio embeddings ${f_{\bm{a}} \in {\mathbb{R}^{L_a \times D{\bm{a}}}}}$ and question features ${f_{\bm{q}} \in {\mathbb{R}^{D_{\bm{q}}}}}$, where $D{\bm{v}}$, $D{\bm{a}}$, and $D{\bm{q}}$ will be uniformly expressed as $D$.

\subsection{Mutual Correlation Module}
In this section, we introduce an important module proposed in this paper, Mutual Correlation Module (MCM). The overall architecture is shown in Fig. \ref{fig:fig3}, which is divided into two steps, the first step is the shallow audio-visual soft correlation establishment process, in which audio-visuals interact with each other without shifting their weights. The second step is encapsulating audio-visual clues inside text objects, where an aggregator is introduced to perform fine-grained integration.

\subsubsection{Association Block}
Images and audio are heterogeneous data, we do not recommend performing information fusion directly using traditional cross-attention, such as the hard-attention paradigm; while it is possible to filter some regions with lower probability, this operation tends to make the learning of the network more difficult to fit. Therefore, we apply a novel soft association mechanism between the visual and audio branches that ensures that the information about itself is fully learned and that attention across modalities can be added to fine-tune the learning parameters dynamically. Specifically, given $N$ Association Blocks (AB) containing self-attention with cross-attention (we called it Bidirectional Inter-attention), we define the self-attention function $\boldsymbol{\varphi}_{\text{self}}(\cdot)$ and the cross-attention function $\boldsymbol{\varphi}_{\text{cross}}(\cdot)$ as:
\begin{equation}\label{eq:stc3}
\boldsymbol{\varphi}_{\text{self}}\left(\textbf{Q},\textbf{K},\textbf{V}\right) = \boldsymbol{\varphi}_{\text{cross}}\left(\textbf{Q},\textbf{K},\textbf{V}\right) = Softmax \left(\frac{{\textbf{Q} \textbf{K}^T }}{{\sqrt d }} \right) \cdot \textbf{V},
\end{equation}
where $\sqrt d$ is the scaling factor, $\textbf{Q} = \textbf{W}^{\textbf{Q}}q$, $\textbf{K} = \textbf{W}^{\textbf{K}}k$ and $\textbf{V} = \textbf{W}^{\textbf{V}}v$, $q$, $k$ and $v$ denote queries, keys, and values, respectively, and $\textbf{W}^{\textbf{Q}}, \textbf{W}^{\textbf{K}}, \textbf{W}^{\textbf{V}} \in \mathbb{R}^{D \times D}$ are the corresponding attention weights, as shown in Fig. \ref{fig:fig3} (a), $\boldsymbol{\varphi}_{\text{cross}}(\cdot)$ outputs the cross-attention results and weights, respectively. In separate AB, the low-level features of audio and video first undergo a self-attention operation, in order to emphasize the audio-visual union, we attempt to add a temporal cross-attention operation to the self-attention, which first takes the cross-modal information through Dimension Compression (DC, $f_j \in \mathbb{R}^{L_j \times D} \rightarrow 
F_j \in \mathbb{R}^{1 \times D},j\in \{\bm{v}, \bm{a} \}$) to get a global temporal query and then performs $\boldsymbol{\varphi}_{\text{cross}}(\cdot)$ to get the global attention weights matrix $\textbf{W}^{cross}_j \in {\mathbb{R}^{D \times D}},j \in \{\bm{v}, \bm{a}\} $ describing the importance of the cross-attention. The specific operations are given as:
\begin{equation}\label{eq:stc2}
f_{\bm{a}2\bm{v}} = \boldsymbol{\varphi}_{\text{self}} \left( f_{\bm{a}}, f_{\bm{a}}, f_{\bm{a}} \right) + f_{\bm{a}} \times \boldsymbol{\varphi}_{\text{cross}}\left(F_{\bm{v}},f_{\bm{a}},f_{\bm{a}}\right) ,
\end{equation}
\begin{equation}\label{eq:stc3}
f_{\bm{v}2\bm{a}} = \boldsymbol{\varphi}_{\text{self}} \left( f_{\bm{v}}, f_{\bm{v}}, f_{\bm{v}} \right) + f_{\bm{v}} \times \boldsymbol{\varphi}_{\text{cross}}\left(F_{\bm{a}},f_{\bm{v}},f_{\bm{v}}\right) ,
\end{equation}
where $\times$ denotes the matrix product, the cross-feature $f_{j-cross},j \in \{\bm{v}, \bm{a}\}$ is obtained from $f_j \times \textbf{W}^{cross}_j$. The fusion steps of Eq. \ref{eq:stc2} and Eq. \ref{eq:stc3} can be interpreted as follows: querying in terms of global temporal allows for the formation of complete audio-visual features, and adding operation avoids cross-attention bias.

Finally, advanced audio-visual embeddings $\hat{f_j},j\in \{\bm{v}, \bm{a}\}$ are obtained through a fully connected feed-forward (FFN) layer, and a batch normalization (BN) layer in a residual way, which can be expressed as:
\begin{equation}\label{eq:stc4}
\hat{f_{\bm{a}}} = f_{\bm{a}} + \text{BN}\left( f_{{\bm{a}}2{\bm{v}}} \right) + \text{BN}\left( \text{FFN}\left( f_{\bm{a}} + \text{BN}\left( f_{{\bm{a}}2{\bm{v}}} \right) \right) \right),
\end{equation}
\begin{equation}\label{eq:stc5}
\hat{f_{\bm{v}}} = f_{\bm{v}} + \text{BN}\left( f_{{\bm{v}}2{\bm{a}}} \right) + \text{BN}\left( \text{FFN}\left( f_{\bm{v}} + \text{BN}\left( f_{{\bm{v}}2{\bm{a}}} \right) \right) \right). 
\end{equation}
Notably, we will retain the cross-feature $f_{j-cross},j \in \{\bm{v}, \bm{a}\}$ and incorporate contrastive learning to measure coordination between audio and visual.

\subsubsection{Aggregator}\label{Audio-Visual Guidance}
Audio and visual establish the correlation via AB, and next, it needs to be transmitted separately to the aggregator that is used to generate distinct discriminative information and then stored inside the question features. However, the cross-domain text fails to localize image or spectrogram features, thereby impairing the fusion effect. Therefore, we propose an elegant mechanism for mutual attention to perceive important information from long video sequences adaptively and to trivialize redundant video and audio segments. As shown in Fig. \ref{fig:fig3} (d), question features $f_{\bm{q}}$ is first performed cross-attention with queries from the audio-visual, which is to avoid a direct query from $f_{\bm{q}}$ would destabilize the original audio-visual features. We define audio-visual queried textual cross-attention as $\boldsymbol{\zeta}_{j-\bm{q}},j \in \{\bm{v}, \bm{a}\}$ and textual queried audio-visual cross-attention as $\boldsymbol{\zeta}_{\bm{q}-j},j \in \{\bm{v}, \bm{a}\}$. Specifically, given an embedding sequence $f_{j} = \left( f_{j\text{-}1}, ..., f_{j\text{-}i}, ..., f_{j\text{-}k} \right), j \in \{\bm{v}, \bm{a}\}$ obtained by randomly sampling $k$ frames, where $f_{j\text{-}i}$ is the individual embedding of frame $i$. Notably, $f_{\bm{q}}$ does not contain the temporal length, thus we applied a DC operation to the temporal dimension for audio-visual to yield $F_{j}, j \in \{\bm{v}, \bm{a}\}$, and a temporal detailed description $\boldsymbol{h}_{j\text{-}i}, j \in \{\bm{v}, \bm{a}\}$ of the $i$-th frame can be expressed as: 
\begin{equation}\label{eq:stc6}
\begin{split}
\overline{f}_{\bm{q}} = f_{\bm{q}} \oplus \boldsymbol{\zeta}_{j-\bm{q}} \left( F_{j},f_{\bm{q}},f_{\bm{q}} \right), \\
\boldsymbol{h}_{j\text{-}i} = ReLU \left(\boldsymbol{\zeta}_{\bm{q}-j} \left( \overline{f}_{\bm{q}} ,f_{j\text{-}i},f_{j\text{-}i} \right)\right),
\end{split}
\end{equation}
where $ReLU$ activation is for preventing overfitting. After computing the scene descriptions $\boldsymbol{h}_j = \left( \boldsymbol{h}_{j\text{-}1}, ..., \boldsymbol{h}_{j\text{-}i}, ..., \boldsymbol{h}_{j\text{-}k} \right), \boldsymbol{h}_{j} \in \mathbb{R}^{k \times D}$ between audio and visual features, we designed a Top-$k$ selection for highlighting the Top-$k$ most concerned clips. Specifically, using $avgpool$ to refine the description of $\boldsymbol{h}_j$, then selecting the Top-$k/2$ fragment features to obtain $\overline{\boldsymbol{h}}_{j} \in \mathbb{R}^{{\frac{k}{2}} \times D}$, finally, the key scene descriptions are served as clues to be aggregated with the question features to obtain a combinatorial question embedding with visual information $\hat{f_{\bm{vq}}}$, and a combinatorial question embedding with audio information $\hat{f_{\bm{aq}}}$, respectively:
\begin{equation}\label{eq:stc9}
\hat{f_{\bm{vq}}} = \sum\nolimits_{\boldsymbol{h} \in \overline{\boldsymbol{h}}_{\bm{v}}} g\left( \boldsymbol{h}, f_{\bm{q}} \right),
\end{equation}
\begin{equation}\label{eq:stc10}
\hat{f_{\bm{aq}}} = \sum\nolimits_{\boldsymbol{h} \in \overline{\boldsymbol{h}}_{\bm{a}}} g\left( \boldsymbol{h}, f_{\bm{q}} \right),
\end{equation}
\begin{equation}\label{eq:stc11}
g\left( \boldsymbol{h}, f_{\bm{q}} \right) = f_{\bm{q}} \oplus \left( \boldsymbol{h} \odot f_{\bm{q}} \right),
\end{equation}
where we use dot product $\odot$ and element-wise add $\oplus$ operations to store the clues among the question features.

The types of sentences, as well as the keywords, contain rich semantic details, in order to focus on the information related to the question from multiple perspectives, we capture the tokens of the sentence, type, and keywords, and then form the text object embedding $\widetilde{f_{\bm{t}}}$, where the concatenated $\widetilde{f_{\bm{t}}}$ requires passing through a multi-features refinement consisting of a fully connected layer ($\mathbb{R}^{3*D} \rightarrow \mathbb{R}^{D}$). Further, to prevent fragmented clues from losing useful information, we propose a simple local element fusion (LEF), which adds the compressed audio-visual features $F_{\bm{v}}$, $F_{\bm{a}}$ to the textual objects and aggregates them to obtain a global result:
\begin{equation}\label{eq:stc12}
\widetilde{f_g} = \widetilde{f_{\bm{t}}} \odot \left(\widetilde{f_{\bm{t}}} \oplus F_{\bm{v}} \oplus F_{\bm{a}}\right),
\end{equation}
then we design a dropout layer to prevent global noise from affecting inference and a fully connected layer ($\mathbb{R}^{D} \rightarrow \mathbb{R}^{D}$) to global features refinement. Subsequently, we added three learnable parameter matrices $\textbf{M}_{\bm{v}\bm{q}}$, $\textbf{M}_{\bm{a}\bm{q}}$, $\textbf{M}_{g} \in \mathbb{R}^{D \times D}$ for $\hat{f_{\bm{vq}}}$, $\hat{f_{\bm{aq}}}$, and $\widetilde{f_g}$ to adaptively learn the optimal fusion weights. Finally, we concatenate $\widetilde{f_g} \times \textbf{M}_{g}$ with $\hat{f_{\bm{vq}}} \times \textbf{M}_{\bm{v}\bm{q}}$ and $\hat{f_{\bm{aq}}} \times \textbf{M}_{\bm{a}\bm{q}}$, and then go through multi-feature refinement ($\mathbb{R}^{3*D} \rightarrow \mathbb{R}^{D}$) to get the combinatorial question embeddings $\hat{f}_q$.

\subsection{Semantic Approximation Module}
In this section, we will introduce contrastive learning for the audio-visual text triad respectively, where the visual-text contrast and audio-text contrast are for the combinatorial question embeddings to be better guided, and the attentional-level visual-audio contrast is for enhanced coordination.

\subsubsection{Audio-Visual Knowledge Distillation} \label{kd}
Under the guidance of the MCM, the text objects were given detailed clues about the audio-visual substrate. Yet visual and audio embeddings remain semantically distant from textual modality, hence requiring knowledge distillation through combinatorial contrastive learning to bridge the cross-modal semantic gap.

Specifically, We adopt the approach of pulling together the positive pairs and pushing away the negative pairs, where the positive pairs include the audio-visual embeddings and combinatorial question embeddings in the same sample, and the negative pairs include the audio-visual embeddings of the other samples extracted from the mini-batch. In addition, we add an extra projection module $\boldsymbol{\Theta} (\cdot)$ that projects the three feature embeddings into the shared latent space. To simply and efficiently compute the contrastive loss for aligning positive pairs in the shared embedding space, we introduce an InfoNCE-based formula to obtain the contrastive loss ${\mathcal{L}_c}^{\bm{\hat{q}}j}, j \in \{\bm{v}, \bm{a}\} $ for a pair of audio-visual embeddings and combinatorial question embeddings as below:
\begin{equation}\label{eq:loss1}
\begin{split}
p^{(i)}_1 = \exp \left( cos\left( \boldsymbol{\Theta} (\hat{f}_q^{(i)}),\boldsymbol{\Theta} (F_j^{(i)}) \right) / \tau \right), \\
{\mathcal{L}_c}^{\bm{\hat{q}}j} =  - \log \left[ {\frac{p^{(i)}_1}{{\sum\nolimits_{k \ne i}^{B_{neg}} {p^{(k)}_1 } + p^{(i)}_1}}} \right],
\end{split}
\end{equation}
where we use the cosine function $cos$ to compute similarity, $\tau$ denotes the temperature, $B_{neg}$ is the number of negative pairs.

\subsubsection{Audio-Visual Contrastive Learning} \label{cl}
Although \cite{Multi-Granularity} proposed a contrast constraint to maintain a consistent focus between visual-text pairs and audio-text pairs, it fails to consider whether coordination between visual and audio is possible. We emphasize that coordinated audio-visual pairs can reduce biases associated with cross-modal attentional strategies, and therefore we introduce contrastive learning to measure the distance between audio-visual pairs. Specifically, given a cross-feature $f_{j-cross}^i,j \in \{\bm{v}, \bm{a}\}$ of $i$-layer in AB, We first fused and averaged the features in the $N$ ABs to get the average representation $m_{j}, j \in \{\bm{v}, \bm{a}\}$:
\begin{equation}\label{eq:loss2-1}
m_{j} = \sum\nolimits_{k = i}^{N} MeanPool\left(f_{j-cross}^i \right)
\end{equation}
where the $MeanPool$ denotes the average pooling operation. Similar to Eq. \ref{eq:loss1}, the contrastive loss ${\mathcal{L}_c}^{\bm{a}\bm{v}}$ can be obtained as below:
\begin{equation}\label{eq:loss2-2}
\begin{split}
p^{(i)}_2 = \exp \left( {{\Vert {m_{\bm{v}}}^{(i)} \Vert}^T}\Vert {m_{\bm{a}}}^{(i)} \Vert / \tau \right), \\
{\mathcal{L}_c}^{\bm{a}\bm{v}} = - \log \left[ {\frac{p^{(i)}_2}{{\sum\nolimits_{k \ne i}^{B_{neg}} {p^{(k)}_2 } + p^{(i)}_2}}} \right].
\end{split}
\end{equation}

\subsection{Prediction Header}
\subsubsection{Answer Prediction}
We discarded the original audio-visual embeddings in the prediction header and used only the combinatorial question embeddings $\hat{f}_q$ storing the audio-visual clues as the basis for inference. The effectiveness of the method is demonstrated in subsequent experiments. Specifically. We built an MLP layer that consists of a linear layer and a softmax activation function, with the output being the probability interval $y \in (0,1)$ of the candidate answers, and the highest probability value corresponding to $argmax (y)$ as the output answer, followed by our application of the cross-entropy function to compute the prediction loss $\mathcal{L}_{answer}$.
\subsubsection{Learning Objective}
During the training period, we sum up the contrastive loss and the prediction loss $\mathcal{L}_{answer}$, the final total loss $\mathcal{L}$ can be written as below:
\begin{equation}\label{eq:loss2}
\mathcal{L} = \mathcal{L}_{answer} + {\mathcal{L}_c}^{\bm{\hat{q}}\bm{v}} + {\mathcal{L}_c}^{\bm{\hat{q}}\bm{a}} + {\lambda}_{\bm{a}\bm{v}} {\mathcal{L}_c}^{\bm{a}\bm{v}}
\end{equation}
where ${\lambda}_{\bm{a}\bm{v}}$ is the ratio corresponding to the audio-visual contrast, the ${\mathcal{L}_c}^{\bm{\hat{q}}\bm{v}}$ and ${\mathcal{L}_c}^{\bm{\hat{q}}\bm{a}}$ denote the distillation objective, the ${\mathcal{L}_c}^{\bm{a}\bm{v}}$ denotes the contrastive objective.

\section{Experiments}
\label{sec:datasets}
Section. \ref{exp1} describes our experimental settings and the two public datasets, Music AVQA and AVQA, where Music AVQA's Q\&A pairs are the most complex and diverse. Section. \ref{exp2} tests the effect of multimodal late fusion on model generalization performance. Paragraph. \ref{exp3} is a comparison experiment with current state-of-the-art methods. Section. \ref{exp4} conducts extensive ablation experiments to demonstrate the effectiveness of each component of our network. Finally, Section. \ref{exp5} will give qualitative experiments to evaluate the effectiveness of our MCD.

\subsection{Datasets and Experimental Settings} \label{exp1}

\subsubsection{Datasets}Table \ref{tab1} shows a detailed description of the two datasets. \textbf{MUSIC-AVQA} dataset is designed to have multiple question-and-answer combinations within the same video, thus its complexity becomes harder with the number of question-and-answer pairs, of which it contains 9,288 videos with 45,867 question-and-answer pairs, Each video contains about 5 question-and-answer pairs on average, and each question involves both visual and auditory sensations. It has as many as 9 types of questions, including audio-focused counting and comparison problems, visual-focused counting and localization problems, and a full range of existential, counting, localization, comparison, and temporal problems.

\begin{table}[]\centering
\caption{Detailed description of Music-AVQA and AVQA datasets. B, O denotes the background sound and object sound, respectively}\label{tab1}
\begin{tabular}{lcccc}
\toprule  
Dataset & \makecell[c]{Video/Q\&A \\ number} & \makecell[c]{Question type \\ number} &\makecell[c]{Sound \\ type}&\makecell[c]{Scene \\ type}\\
\midrule  
AVQA &  57.0K/57.3K & 8 & B & Real-life\\
Music-AVQA &  9.3K/45.9K & 9 & O & Music\\

\bottomrule  
\end{tabular}
\end{table}

\textbf{AVQA} has a high video count of 57,015 and contains 57,335 question-and-answer pairs. Unlike Music-AVQA, it is multiple-choice, i.e., choose the most appropriate of four answers, and most quiz scenarios are in real-world environments. Similarly, AVQA provides visual and audio information, and 8 types of questions.

\subsubsection{Experimental Settings}
Regarding the basic encoders, for Music AVQA, we tested three backbone networks on the visual branch, including ResNet, CLIP, and SwinV2 \cite{swin}, whose output visual feature dimensions are unified as ${f_{\bm{v}} \in {\mathbb{R}^{L_v \times 512}}}$. On the audio branch, we used the features ${f_{\bm{a}} \in {\mathbb{R}^{L_a \times 512}}}$ extracted by the officially recommended Vggish network. On AVQA, we uniformly used the officially published visual backbone network resnext101 with the features ${f_{\bm{v}} \in {\mathbb{R}^{L_v \times 2048}}}$ and ${f_{\bm{a}} \in {\mathbb{R}^{L_a \times 2048}}}$ extracted by the audio backbone network PANNs \cite{PANNs}, followed by an MLP layer $\left(\mathbb{R}^{L \times 2048} \rightarrow \mathbb{R}^{L \times 512} \right)$ to get the final feature embedding. For both of the above datasets, their input textual information is embedded by LSTM \cite{lstm}, where the number of hidden units is consistent with the audio-visual dimension $\mathbb{R}^{512}$.

During the training period, the batch size is defaulted by 64, our MCD uses two NVIDIA A5000 GPUs and uses the Adam optimizer with an initial learning rate of $0.001$ uniformly on both datasets, followed by multiplying a factor of $0.1$ every $8$ rounds to decrease the learning rates for a total of 30 epochs. For audio-visual contrastive loss ratio ${\lambda}_{\bm{a}\bm{v}}$ is set to 0.1 by default, which is to avoid the weighting of the audio-visual contrast auxiliary task is too large to the inference.

Notably, we attempted to train on the original image set and audio set obtained by sampling 1 frame every 6 seconds on the video, but the computational resource limitations made the training time too long and did not yield the desired results, so we discarded the end-to-end reasoning strategy.

\subsection{Impact of Decision-Level Fusion on Inference} \label{exp2}
Current approaches assume that deep audio-visual features added to the prediction can help the network better infer the most appropriate answer, yet ignoring the situation where multiple questions may coexist in a single scenario. To explore whether the late multimodal fusion affects the accuracy of the inference engine, we first perform a simple fusion test on Music-AVQA and exhibit the results in Table \ref{tab2}, the results of unimodal inference can be intuitively seen that the performance of visual alone is the best. For the ablation test of multimodal fusion, we design a multimodal MLP layer, and the visual, audio, and question embeddings from basic encoders are concatenated to simultaneously input to the MLP layer to obtain the logits distribution. After observing the average accuracy of the predictions, we can find that the increase in the number of modality combinations brings no gain to the inference, but rather a decrease.

\begin{figure}
\centering
\subfigure[Our main idea of removing audio-visual fusion.] {
\includegraphics[scale=0.55]{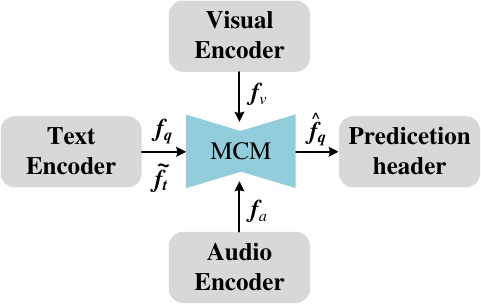}
}

\subfigure[Methods for testing late-fusion.] { 
\includegraphics[scale=0.55]{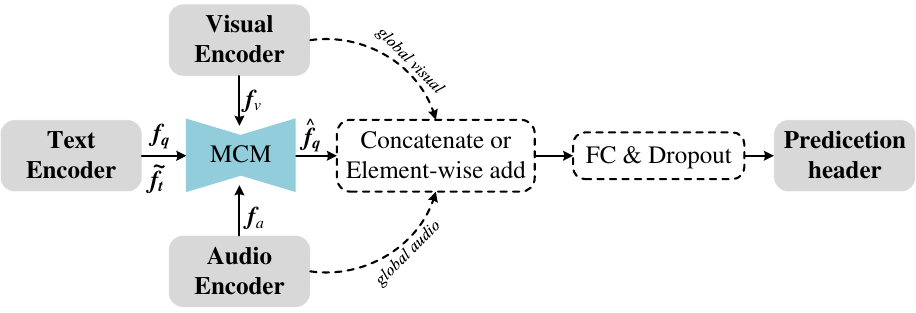}
}    

\caption{Our proposed method of testing multimodal fusion, (a) demonstrates the abandonment of decision-level fusion, (b) demonstrates the typical concatenation and element-wise add process.}
\label{fig:exp1} 
\end{figure}

\begin{figure}
\centering
\subfigure[The inference without visual and audio modalities.] {
\includegraphics[scale=0.4]{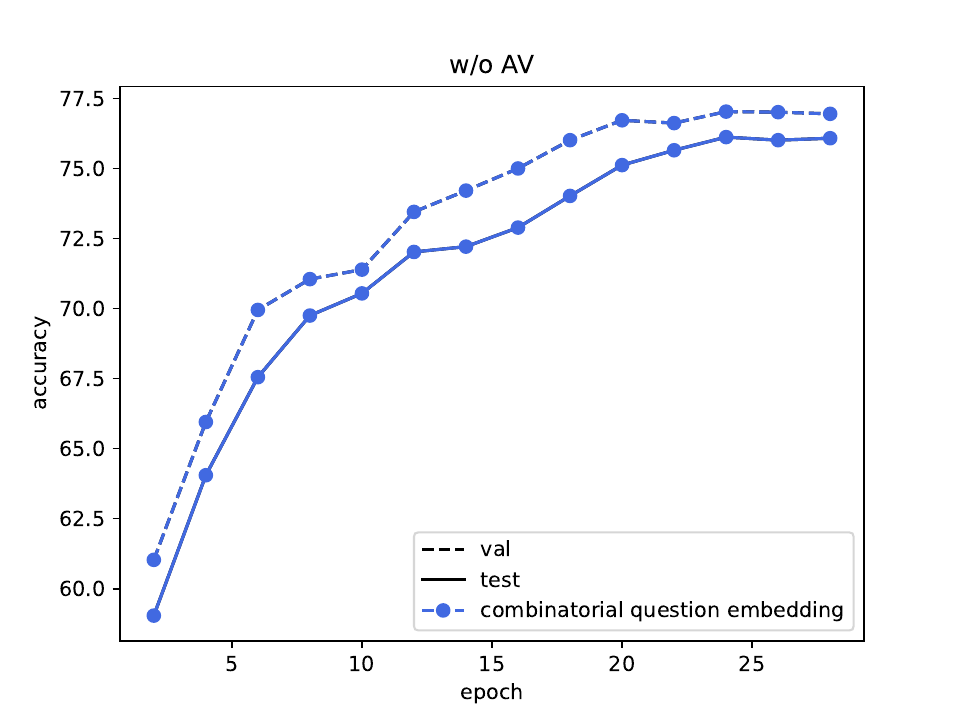}
}

\subfigure[The inference with audio-visual late-fusion.] { 
\includegraphics[scale=0.35]{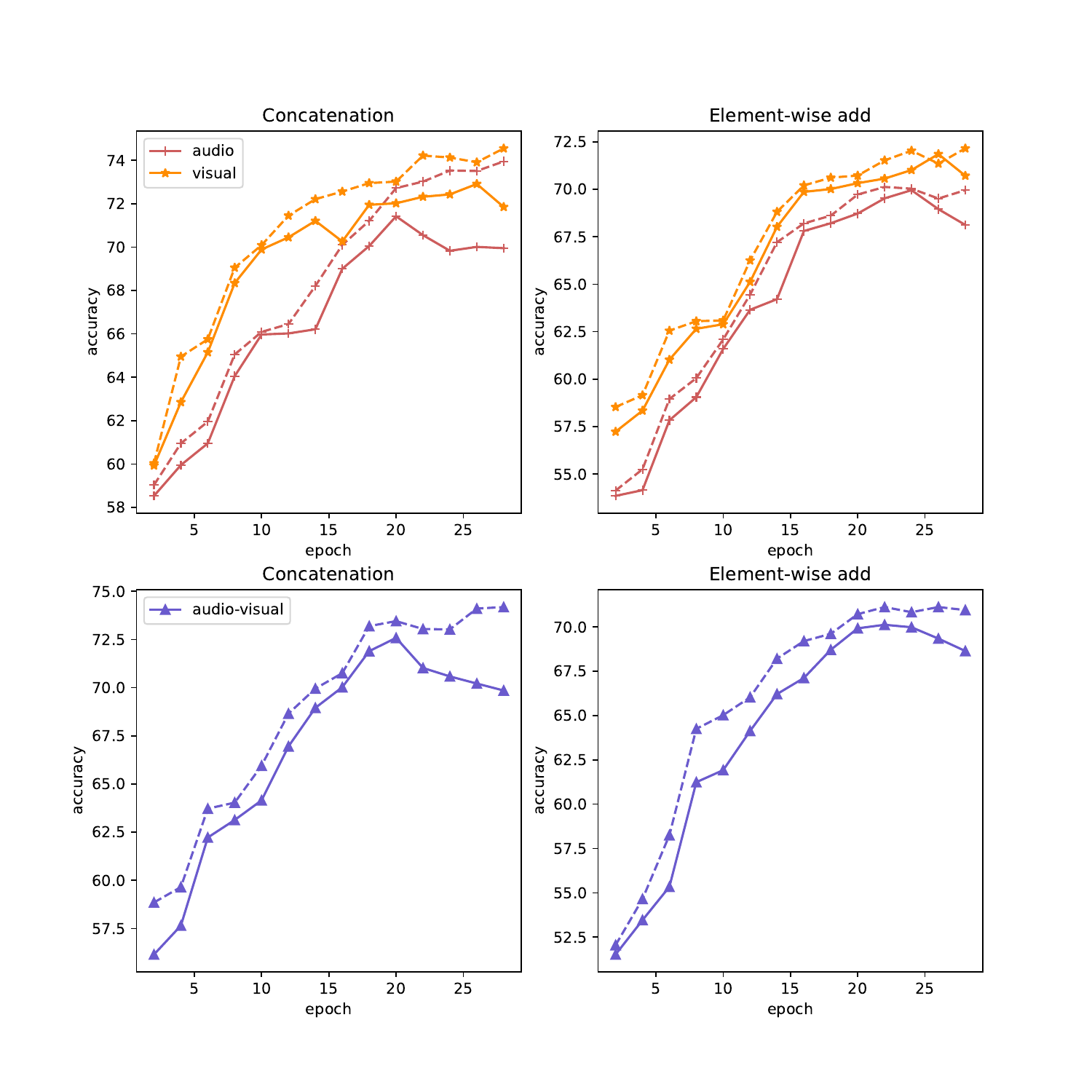}
}    

\caption{An illustration of the impact of multimodal late fusion on inference results.}
\label{fig:exp2} 
\end{figure}

For further testing, we add two typical decision-level fusion methods to the proposed MCM. As shown in Fig. \ref{fig:exp1}, after obtaining the combinatorial question embeddings of MCM, (b) demonstrating the process of element-wise addition or concatenation with the global audio-visual embeddings, in which we add to both methods a full-connection layer for feature refinement, and a dropout layer to prevent overfitting. After testing on the Music-AVQA dataset, the results are shown in Fig. \ref{fig:exp2}(b), when fusion is performed using concatenation, the first figure shows the inference with joining audio-only embedding or visual-only embedding is fitted around the 20th epoch, and with further training, the network starts to rely on either audio or visual redundant parameters, the validation set results start to drift while the test-set results start to grow negatively, where the most pronounced negative effect is the addition of visual-only. The figure in the lower left corner shows the network with both audio-visual embeddings has started to fit around the 15th epoch, with further training, the validation set results continue to rise while the test set results start to grow substantially negatively, which further reveals that audio-visual embeddings are not friendly to the inference engine but instead impair the network's ability to generalize. Similarly, the figures in the second column show the results when comparing the adoption of elemental fusion, although the addition of the audio-visual embedding elements does not show the large negative growth that occurs in concatenation, there is no significant increase in accuracy, but rather a decrease. On the contrary, as shown in Fig. \ref{fig:exp2}(a), our idea of discarding fusion is confirmed, as the performance of the validation and test sets is stable around the 25th epoch and does not show a large drop.

\begin{table}
\centering
\caption{Experimental results(\%) of different input modality combinations on Music-AVQA. \textbf{Q}, \textbf{A}, and \textbf{V} denote the question, audio, and visual modality. Variations 1 and 2 denote changes relative to \textbf{Q}+\textbf{A} and \textbf{Q}+\textbf{V}, respectively. The table results show that the increase in the number of modality combinations fails to improve the performance.}\label{tab2}
\begin{tabular}{|ccc|c|cc|}
\hline
\multicolumn{3}{|c|}{Modality} & \multirow{2}{*}{\makecell[c]{Overall \\ Accuracy}} & \multicolumn{2}{c|}{\multirow{2}{*}{Variations}}\\
\cline{1-3}
\textbf{Q} & \textbf{A} & \textbf{V} & & &\\
\cline{1-3} \cline{4-6}
\checkmark & & & 51.24 & \multicolumn{2}{c|}{-}\\
 & \checkmark & & 54.60& \multicolumn{2}{c|}{\textcolor{blue}{\textbf{($\uparrow$ 3.36)}}}\\
 & & \checkmark & 61.98 & \multicolumn{2}{c|}{\textcolor{blue}{\textbf{($\uparrow$ 10.74)}}}\\
\hline
\checkmark & \checkmark & & 67.26 & \multicolumn{2}{c|}{-}\\
\checkmark &  & \checkmark & 68.31 & \multicolumn{2}{c|}{-}\\
\checkmark & \checkmark & \checkmark & 66.72 & \textcolor{red}{\textbf{($\downarrow$ 0.54)$^{1}$}} &\textcolor{red}{\textbf{($\downarrow$ 1.59)$^{2}$}} \\
\hline
\end{tabular}
\end{table}

\subsection{Comparison With the State-of-The-Art Methods} \label{exp3}
To highlight the strengths of our network, we compared current state-of-the-art methods on two publicly available datasets. 

\subsubsection{Results on the Music-AVQA}
Table \ref{MUSIC-AVQA} shows the results of comparing our MCD with other state-of-the-art methods, of which there are 13 evaluations: for the test results focusing on \textbf{Audio} and \textbf{Visual} type, our method completely outperforms the previous methods due to the soft association mechanism in MCD which can prevent information loss, and in terms of generalizability, our approach achieves relatively good results for different problem types. In terms of overall accuracy, our method currently outperforms other state-of-the-art methods by 2.6 \%. Further, we analyze that:
\begin{itemize}
\item \textbf{AVST} achieves good inference as a baseline, but in the actual experiments, the model has been largely fitted around 10$\sim$12th epochs of training and the accuracy of the validation set starts to flotation. For this phenomenon, although the network produced effective interactions on audio-visual in the middle layer, the computational parameters that come with bias the network towards single scene information, in other words, when a single video has multiple question-answer pairs, the late fusion not only won't help the network to inference, the Instead, it will reduce the generalization ability of the network.
\item \textbf{PSTP-Net} is relatively less computationally intensive due to the selection of key audio-visual clips as inputs. However, the importance of a comprehensive understanding of audio has been neglected in practical applications, leading to poor question reasoning in \textbf{Audio} types.
\end{itemize}

\begin{table*}[]\centering
\caption{Results(\%) of different methods on the test set of MUSIC-AVQA, where question types include counting, comparative, localization, existential, and temporal. AVG denotes the average accuracy for corresponding scenarios, OA denotes the overall accuracy}\label{MUSIC-AVQA}
\begin{tabular}{l|lll|lll|llllll|c}
\hline
\multirow{2}{*}{Method}  & \multicolumn{3}{c|}{\textbf{Audio}} & \multicolumn{3}{c|}{\textbf{Visual}} & \multicolumn{6}{c|}{\textbf{Audio-Visual}} & \multirow{2}{*}{\textbf{OA}} \\
      &   Count    &   Comp    &  \textbf{Avg.}    &   Count   &   Local    & \textbf{Avg.}    & Exist  & Count  & Local  &  Comp & Temp  & \textbf{Avg.}  &   \\
\hline
FCNLSTM \cite{FCNLSTM}     &  70.80& 65.66& 68.90& 64.58& 48.08& 56.23& 82.29& 59.92& 46.20& 62.94& 47.45& 60.42& 60.81 \\
GRU \cite{GRU}   &   71.29& 63.13& 68.28& 66.08& 68.08& 67.09& 80.67& 61.03& 51.74& 62.85& 57.79& 63.03& 65.03\\
HCAttn \cite{HCAttn}  &  70.80& 54.71& 64.87& 63.49& 67.10& 65.32& 79.48& 59.84& 48.80& 56.31& 56.33& 60.32& 62.45 \\
MCAN \cite{MCAN}   & 78.07& 57.74& 70.58& 71.76& 71.76& 71.76& 80.77& 65.22& 54.57& 56.77& 46.84& 61.52& 65.83 \\
PSAC \cite{PSAC}    &  75.02& 66.84& 72.00& 68.00& 70.78& 69.41& 79.76& 61.66& 55.22& 61.13& 59.85& 63.60& 66.62  \\
HME \cite{HME}    & 73.65& 63.74& 69.89& 67.42& 70.20& 68.83& 80.87& 63.64& 54.89& 63.03& 60.58& 64.78& 66.75  \\
HCRN \cite{HCRN}   & 71.29& 50.67& 63.69& 65.33& 64.98& 65.15& 54.15& 53.28& 41.74& 51.04& 46.72& 49.82& 56.34 \\
AVSD \cite{AVSD}    &  72.47& 62.46& 68.78& 66.00& 74.53& 70.31& 80.77& 64.03& 57.93& 62.85& 61.07& 65.44& 67.32  \\
Panp-AVQA \cite{pianoavqa}  &  75.71& 65.99& 72.13& 70.51& 75.76& 73.16& 82.09& 65.38& 61.30& 63.67& 62.04& 66.97& 69.53  \\
AVST \cite{musicavqa}   &  77.78& 67.17& 73.87& 73.52& 75.27& 74.40& 82.49& 69.88& 64.24& 64.67& 65.82& 69.53& 71.59  \\
PSTP-Net \cite{pstp-net}   &  73.97& 65.59& 70.91& 77.15& 77.36& 77.26& 76.18& 73.23& \textbf{71.80}& 71.79& \textbf{69.00}& 72.57& 73.52 \\
\hline
MCD \textbf{(Ours)}  &  \textbf{82.40}  & \textbf{67.39}  & \textbf{77.31} & \textbf{80.45} &\textbf{81.20}&\textbf{81.06} &\textbf{83.83} &\textbf{75.81} &70.63 &\textbf{72.42} &67.42&\textbf{75.51}& \textbf{76.12}\\
\hline
\end{tabular}
\end{table*}

\begin{table*}[]
\centering
\caption{Comparison results(\%) of different visual backbones and methods on Music-AVQA. $\triangle$ indicates the relative improvement of the same pre-trained model compared to other methods. a-avg,v-avg, and av-avg denote the average accuracy regarding audio, visual, and audiovisual question types, respectively.}\label{backbone}
\begin{tabular}{cccccccccc}
\toprule  
Method & \makecell[c]{Pre-trained \\ model}& A-avg & $\triangle$ $\uparrow$ &V-avg & $\triangle$ $\uparrow$ & AV-avg & $\triangle$ $\uparrow$ & OA & $\triangle$ $\uparrow$\\
\midrule  
AVST \cite{musicavqa}   & ResNet-18 & 73.87 &-& 74.40 &- & 69.53 &- & 71.59 &-\\

\textbf{Ours}& ResNet-18 & 76.41 & \textcolor{blue}{\textbf{($\uparrow$ 2.54)}}  & 79.73 & \textcolor{blue}{\textbf{($\uparrow$ 5.33)}} & 71.66 & \textcolor{blue}{\textbf{($\uparrow$ 2.13)}} & 74.64& \textcolor{blue}{\textbf{($\uparrow$ 3.05)}}\\
\midrule  
PSTP-Net \cite{pstp-net} & CLIP-ViT-B/32 & 70.91 &-& 77.26 &-& 72.57 &-& 73.52 &-\\

\textbf{Ours} & CLIP-ViT-B/32 & 76.29 &\textcolor{blue}{\textbf{($\uparrow$ 5.38)}} & 80.31 &\textcolor{blue}{\textbf{($\uparrow$ 3.05)}} & 73.37 &\textcolor{blue}{\textbf{($\uparrow$ 0.80)}} & 75.42&\textcolor{blue}{\textbf{($\uparrow$ 1.90)}}\\
\midrule  
LAVisH  \cite{lavish} & Swin-V2-L & 75.49 &-& 76.41 &-& 75.12 &-& 74.61 &-\\

\textbf{Ours}  & Swin-V2-L & \textbf{77.31} &\textcolor{blue}{\textbf{($\uparrow$ 1.82)}} & \textbf{81.06} &\textcolor{blue}{\textbf{($\uparrow$ 4.65)}} & \textbf{75.51} &\textcolor{blue}{\textbf{($\uparrow$ 0.39)}} & \textbf{76.12} & \textcolor{blue}{\textbf{($\uparrow$ 1.51)}}\\
\bottomrule  
\end{tabular}
\end{table*}

\subsubsection{Results on the AVQA}
Yang \textit{et al.} \cite{avqa} proposes a hierarchical audio-visual fusion module adaptable to arbitrary networks that brings enhancements to the original network. To obtain more convincing results, we apply MCM and SA on the encoder of the state-of-the-art method and perform tests on the AVQA dataset. Notably, we additionally input tokens of answer options to form text objects. The results are shown in Table \ref{AVQA-res}, where our model achieves the best results, and despite we have no significant improvement in scenarios with a relatively limited number of question-answer pairs, the proposed MCD still has significant potential for reasoning in audio-visual question-answering tasks.

\begin{table}[]\centering
\caption{Results(\%) of different methods on the test set of AVQA. Avg denotes the average accuracy across all question types.}\label{AVQA-res}
\begin{tabular}{lccccccccc}
\toprule  
Method & \makecell[c]{Ensemble \\ HAVF \cite{avqa}}&  Avg \\
\midrule  
HME \cite{HME}& \checkmark &  85.0   \\
PSAC \cite{PSAC} & \checkmark & 87.4   \\
LADNet \cite{LADNet} & \checkmark & 84.1   \\
ACRTransformer \cite{ACRTransformer} & \checkmark & 87.8   \\
HGA \cite{HGA} & \checkmark & 87.7   \\
HCRN \cite{HCRN} & \checkmark & 89.0   \\
\midrule  
PSTP-Net \cite{pstp-net} & $\times$ & 90.2   \\
\midrule  
MCM w/o SA \textbf{(Ours)}  & $\times$ & 89.5   \\
MCM w/ SA \textbf{(Ours)}  & $\times$ & \textbf{90.8}   \\
\bottomrule  
\end{tabular}
\end{table}

\subsection{Ablation Study}  \label{exp4}

\subsubsection{Basic Encoders Replacement Experiments}
To demonstrate the strong generalization of MCD for different backbones, we conducted extensive replacement experiments on the Music-AVQA dataset.

\textbf{Visual Encoder.} We tested three of the currently most popular visual networks on the Music-AVQA dataset. Table \ref{backbone} shows the overall test results of our model against other state-of-the-art methods under the same pretraining backbone. Intuitively, MCD outperforms the compared models comprehensively, where the difference in results is most pronounced for AVST using ResNet and PSTP-Net using CLIP. It is worth noting that LAVisH proposes to adaptively perceive the audio-visual streams from the raw data, which is extremely computationally intensive in terms of running time and computation. Due to machine performance constraints, we had to change the original method of sampling 10 frames for each video to 7 frames for each video, and the batch size was changed to 8 instead of the original 16.

\textbf{Text Encoder.} For question information, we briefly tested the effect of LSTM as well as Bi-LSTM \cite{bilstm} on network inference. As shown in Table \ref{lstm}, the substitution of the feature extraction network in the input layer does not affect the results due to the small number of problem texts. 

Notably, considering that audio information does not have a large number of complex and fine-grained features, as well as the official Vggish network employed on Music-AVQA and the PANNs network employed on AVQA are sufficient to extract the spectrogram features, no pre-training network substitution experiments for audio are conducted in this paper.

\begin{table}
\centering
\caption{Results(\%) of different text encoder comparisons and text objects ablation on Music-Avqa. \textbf{T} denotes the text objects in this paper.}\label{lstm}
\begin{tabular}{c|c|ccc}
\hline
\multirow{2}{*}{Text encoder}  & \multirow{2}{*}{\textbf{T}} & \multicolumn{3}{c}{Music-AVQA} \\
\cline{3-5}
& & A-avg & V-avg & AV-avg\\
\hline
\multirow{2}{*}{LSTM}   &   $\times$      & 74.68 & 79.12 & 72.18 \\ 
    &   \checkmark  &  \textbf{77.31} & 81.06 & \textbf{75.51}\\ 
\hline
 \multirow{2}{*}{Bi-LSTM}     &   $\times$      & 74.12 & 80.01 & 72.95\\ 
    &   \checkmark  &   76.84 & \textbf{81.12} & 74.98\\ 
\hline
\end{tabular}
\end{table}

\subsubsection{Imapct of Different Component}
To validate the necessity of each component in the network, we first tested the different component variants on Music-AVQA and AVQA datasets. 

\textbf{Bidirectional Inter-attention.} As shown in Table \ref{AB-att}, we can observe that there is a relatively small improvement in the results of self-attention manipulation at the feature level compared to the results of reasoning using only feature embedding. Secondly, when using cross-attention to generate interactions, some audio-related visual regions or visually related audio clips were attended to, resulting in a relatively large uptick compared to the results with self-attention. The Bidirectional Inter-attention proposed in this paper, which establishes soft associations across modalities and mitigates information loss, bridging the hard attention drawbacks, gives us an advantage in inference when comparing the above two approaches. 

\textbf{Number of Association Blocks.} As shown in Table \ref{FLOPs}, we tested the effect of different numbers of ABs on the inference as well as the network parameters, PSTP-Net benefits from cropping fewer images on the raw video is computationally stronger than MCD, but collectively, our performance is superior when the number of parameters as well as the computational speed are not much different.

\textbf{Local Element Fusion.} To explore whether LEF contributes to the aggregator's ability to retrieve the lost audiovisual information, we perform ablation tests for Eq. \ref{eq:stc12} on Table \ref{ele-test}, which illustrates that when removing either the global visual elements or the audio elements reduces the desirable results, and when removing the utilization of the textual objects for fusion with the global audio-visual, the network regresses to its original dependence on the audiovisual, resulting in performance degradation. Notably, we emulate the operation of storing clues ($\oplus$ and $\odot$), which performs better than using only element-wise add.

\textbf{Knowledge Distillation and Contrastive Learning.} Our main idea is to learn the latent space of approximating cross-modal gaps in order to indirectly transfer modal knowledge. Table \ref{sa-test} demonstrates the test effect of semantic approximation during the training process, and we observe that: for the distillation losses ${\mathcal{L}_c}^{\bm{\hat{q}}\bm{v}}$ and ${\mathcal{L}_c}^{\bm{\hat{q}}\bm{a}}$, our SA has a great potential to mitigate semantic ambiguities, and the ablation results confirm that knowledge distillation achieve a relative improvement for AVQA task. For contrasting coordination on audio-visual, although it does not lead to greater gains, learning the original audio-visual federation under the contrast goal produces more stable reasoning results.

\begin{table}
\centering
\caption{Result Comparison(\%) of different attention methods in AB. V-query and A-query denote cross-attention using visual or audio as query, respectively, and Bidirectional Inter-attention is proposed in this paper.}\label{AB-att}
\begin{tabular}{c|cccc|c}
\hline
\multirow{2}{*}{Method} & \multicolumn{4}{c|}{Mode} & \multirow{2}{*}{OA} \\
 & Audio & Visual & A-query & V-query & \\
\hline
w/o attention &-& - &- & - &66.00 \\
\hline
 & \checkmark &  &  &  & 65.45 \\
Self-attention    &  & \checkmark  &  &  & 67.34 \\
   & \checkmark & \checkmark &  &  & 68.24 \\
\hline
  &  &  & \checkmark  &  & 69.12 \\
Cross-attention   &  &   &  & \checkmark & 70.34 \\
   &  &  & \checkmark & \checkmark & 71.35 \\
\hline
 Bidirectional & \checkmark & \checkmark  & \checkmark  &  & 72.04 \\
Inter-attention   & \checkmark  & \checkmark  &  & \checkmark & 74.54 \\
   & \checkmark & \checkmark & \checkmark & \checkmark & \textbf{76.12} \\
\hline
\end{tabular}
\end{table}
\begin{table}[]
\centering
\caption{Parameter and FLOPs of competitive approaches and ours.}\label{FLOPs}
\begin{tabular}{cccc}
\toprule  
Method & Training Param (M) & FLOPs (G) & OA ($\%$) \\
\midrule  
AVST \cite{musicavqa}   & 18.480 &3.188 & 71.59 \\
PSTP-Net \cite{pstp-net}  & 4.297& 1.223& 73.52\\
\midrule  
\textbf{AB} (N=1)  &  6.026 & 2.173 & 76.01 \\
\textbf{AB} (N=2)  &  7.587 & 2.542 & \textbf{76.12} \\
\textbf{AB} (N=3)  &  9.148 & 2.850 & 75.98 \\
\bottomrule  
\end{tabular}
\end{table}
\begin{table}
\caption{Ablation test results(\%) on element fusion (E\lowercase{q}. \ref{eq:stc12}). $F_{\bm{v}}$ and $F_{\bm{a}}$ denote the global temporal visual and audio embeddings, and $\widetilde{f_{\bm{t}}}$ denotes the text object embedding.}\label{ele-test}
\centering
\begin{tabular}{ccccc}
\toprule
\multirow{2}{*}{Modality} & \multicolumn{4}{c}{Music-AVQA}\\ 
\cmidrule(lr){2-5}
      &     A-avg     &   V-avg   &    AV-avg    &    OA  \\ 
\midrule
w/o $F_{\bm{v}}$   &     76.05     &      79.92    &  72.62  &   72.22  \\
w/o $F_{\bm{a}}$   &     75.87     &      80.06    &  74.57  &   74.82 \\
w/o $\widetilde{f_{\bm{t}}}$   &    74.68      &    79.12     &  72.18  &  73.12  \\
\midrule
\multicolumn{5}{c}{\textbf{Element Fusion Methods}} \\
\midrule
Only element-wise add   &      76.28    &     80.54     &    74.41   & 75.08 \\ 
\rowcolor{green!10} w/ dot-product   &      77.31    &     81.06     &    75.51   & 76.12 \\ 
\bottomrule
\end{tabular}
\end{table}
\begin{table}
\caption{Ablation test results(\%) on contrastive learning and knowledge distillation. ${\mathcal{L}_c}^{\bm{a}\bm{v}}$ denotes the audio-visual contrastive loss, and ${\mathcal{L}_c}^{\bm{\hat{q}}\bm{v}}$ and ${\mathcal{L}_c}^{\bm{\hat{q}}\bm{a}}$ in the semantic approximation method denote the distillation loss of visual and audio, respectively.}\label{sa-test}
\centering
\begin{tabular}{ccccccc}
\toprule
\multirow{2}{*}{Method} & \multicolumn{4}{c}{Music-AVQA} & & AVQA  \\ 
\cmidrule(lr){2-5} \cmidrule(lr){7-7}
      &     A-avg     &   V-avg   &    AV-avg    &    OA   &   & Avg \\ \midrule
MCD w/o ${\mathcal{L}_c}^{\bm{\hat{q}}\bm{v}}$   &     76.25     &      79.52    &  74.82  &   74.95 &  & 90.14 \\
MCD w/o ${\mathcal{L}_c}^{\bm{\hat{q}}\bm{a}}$   &     74.45     &      81.05    &  74.07  &   74.52 &  & 90.32 \\
MCD w/o ${\mathcal{L}_c}^{\bm{a}\bm{v}}$   &    76.87      &     81.02     &  74.92  &  75.63  &  & 90.20 \\\midrule
\rowcolor{green!10} MCD   &      77.31    &     81.06     &    75.51   & 76.12 & & 90.78 \\ 
\bottomrule
\end{tabular}
\end{table}

\subsection{Qualitative analysis} \label{exp5}
\begin{figure*}
	\centering
		\includegraphics[scale=0.3]{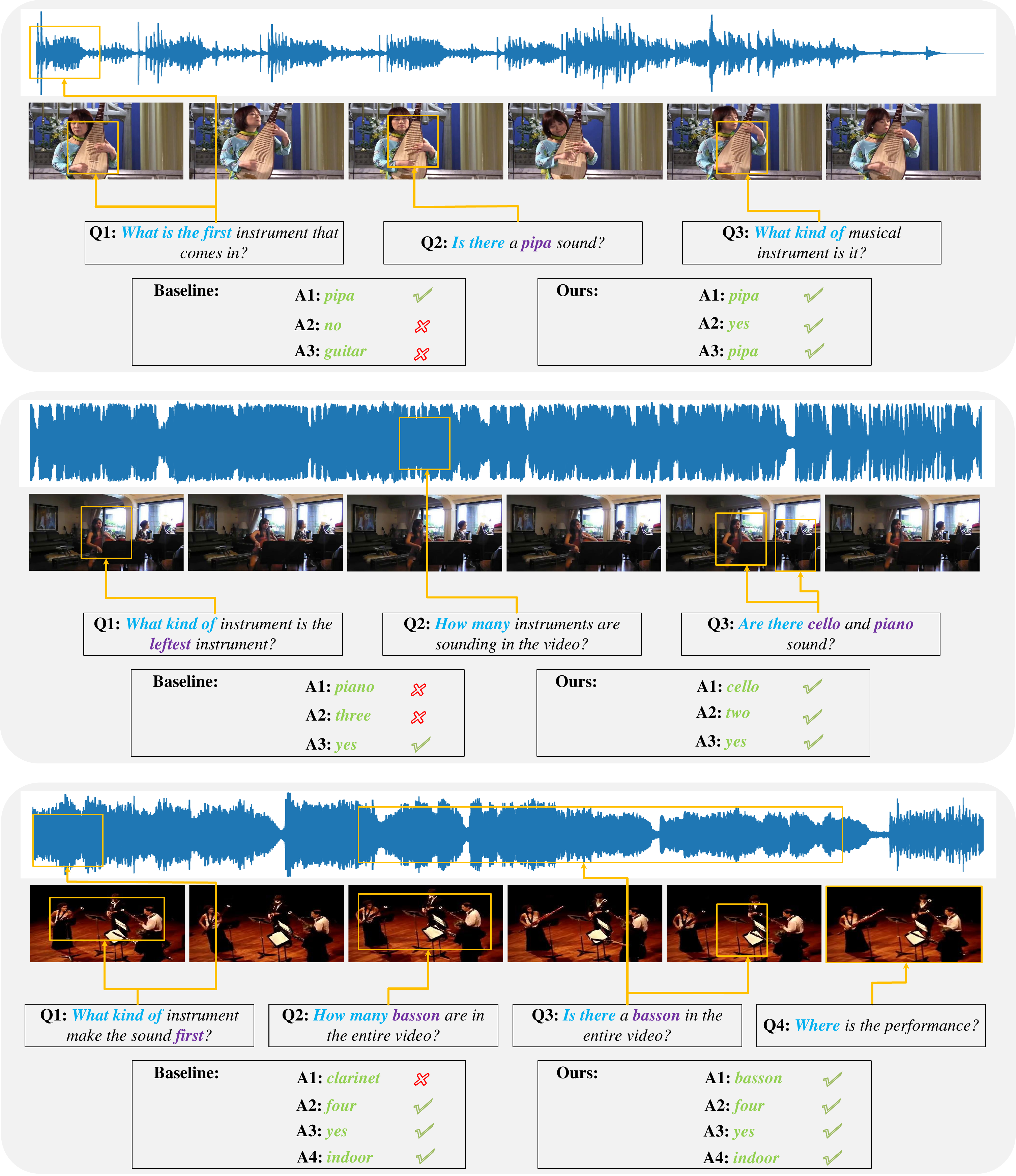}
	\caption{Three qualitative results on the Music-AVQA test set and comparison with the baseline model. Each video contains three to five questions, which include question types such as object localization, counting, and existence. The purple color is the fine-grained semantic information that our method focuses on, orange lines guided the visual areas and audio clips associated with the questions.}
	\label{fig:exp-final}
\end{figure*}

To highlight the superiority of the proposed MCD in multi-question-answer pairs scenarios, we compare it with the baseline model AVQA and present the multi-question reasoning results of three instances on Music-AVQA in Fig. \ref{fig:exp-final}. Four types of questions, ``object localization", ``existence", ``counting" and ``scene localization", appear in the three instances. As a whole, MCD is capable of accurately locating the type of question and selecting the most correct answer due to its question-driven reasoning approach.

For example, \textbf{Q3} of instance 1, \textbf{Q1} of instance 2, and \textbf{Q1} of instance 3 are all questions about ``\textit{what kind of}", which require the inference engine to have the ability to recognize the scene information and understand the unique question. However, the current approach overly relies on audio-visual information and lacks the ability to understand the details of the problem. When reasoning, e.g., in \textbf{Q1} of instance 2, even if ``\textit{piano}" exists in the scenario, the baseline model is prone to make a wrong choice by ignoring ``\textit{leftest}". On the contrary, our model is able to locate the fine-grained semantic information and derive the correct answer ``\textit{cello}" by using the question features with a small quantity of audio-visual clues.

\section{Conclusion and limitation}
\textbf{Conclusion.} In this paper, we delve into the solution of complex question-answer scenarios, simulating human thinking to explore how audio-visual features can be utilized to aid question derivation. We first design an association block, which cleverly uses residual structural connections to enhance information transfer across modal attention on top of self-attention. Subsequently, we designed an aggregator to hierarchically write semantic information describing the details of scenes into the question features, allowing the questions to be attached with key clues. We also investigated a contrast-based knowledge distillation method to approximate the semantic gap between questions and audiovisuals, and coordinated the balance between audio-visual through contrastive learning. Extensive experiments show that our model outperforms previous state-of-the-art methods and is particularly suitable for tasks with diverse question types.

\textbf{Limitation.} However, our MCD relies on powerful pre-trained encoders, a move made to minimize the huge amount of computation associated with raw data. In the future, MCD will be extended to question-answering tasks in unimodal scenarios.

\end{document}